\documentclass[letterpaper]{article} 
\usepackage{aaai25}  
\usepackage{times}  
\usepackage{helvet}  
\usepackage{courier}  
\usepackage[hyphens]{url}  
\usepackage{graphicx} 
\usepackage{natbib}  
\usepackage{caption} 
\usepackage{algorithm}
\usepackage{algorithmic}

\usepackage{newfloat}
\usepackage{listings}
\usepackage{multirow}
\usepackage{amssymb}
\usepackage{amsmath}
\usepackage{cases}
\usepackage{subfig}
\usepackage[capitalize]{cleveref}
\usepackage{booktabs}
\DeclareCaptionStyle{ruled}{labelfont=normalfont,labelsep=colon,strut=off} 
\floatstyle{ruled}
\newfloat{listing}{tb}{lst}{}
\floatname{listing}{Listing}
\nocopyright 

\title{GeoBEV: Learning Geometric BEV Representation \\ for Multi-view 3D Object Detection}
\author{
    Jinqing Zhang\textsuperscript{\rm 1}, Yanan Zhang\textsuperscript{\rm 1}, Yunlong Qi\textsuperscript{\rm 2}, Zehua Fu\textsuperscript{\rm 3}, Qingjie Liu\textsuperscript{\rm 1,3,4}\thanks{Corresponding author.}, Yunhong Wang\textsuperscript{\rm 1,3}\\
}
\affiliations{
    \textsuperscript{\rm 1}State Key Laboratory of Virtual Reality Technology and Systems, Beihang University, Beijing, China\\
    \textsuperscript{\rm 2}Beijing Jingwei Hirain Technologies Co., Inc. \\
    \textsuperscript{\rm 3}Hangzhou Innovation Institute, Beihang University, Hangzhou, China\\
    \textsuperscript{\rm 4}Zhongguancun Laboratory, Beijing, China\\
    \tt \{zhangjinqing, zhangyanan\}@buaa.edu.cn, yunlong.qi@hirain.com, \\ \{zehua\_fu, qingjie.liu, yhwang\}@buaa.edu.cn
}

\usepackage{bibentry}

\begin{document}

\maketitle

\begin{abstract}

Bird's-Eye-View (BEV) representation has emerged as a mainstream paradigm for multi-view 3D object detection, demonstrating impressive perceptual capabilities. However, existing methods overlook the geometric quality of BEV representation, leaving it in a low-resolution state and failing to restore the authentic geometric information of the scene. In this paper, we identify the drawbacks of previous approaches that limit the geometric quality of BEV representation and propose Radial-Cartesian BEV Sampling (RC-Sampling), which outperforms other feature transformation methods in efficiently generating high-resolution dense BEV representation to restore fine-grained geometric information. Additionally, we design a novel In-Box Label to substitute the traditional depth label generated from the LiDAR points. This label reflects the actual geometric structure of objects rather than just their surfaces, injecting real-world geometric information into the BEV representation. In conjunction with the In-Box Label, Centroid-Aware Inner Loss (CAI Loss) is developed to capture the inner geometric structure of objects. Finally, we integrate the aforementioned modules into a novel multi-view 3D object detector, dubbed GeoBEV, which achieves a state-of-the-art result of 66.2\% NDS on the nuScenes test set. \textit{The code is available at https://github.com/mengtan00/GeoBEV.git.}

\end{abstract}

\section{Introduction}

Multi-view 3D object detection stands as a prominent perception paradigm for cost-effective autonomous driving. Presently, many camera-only detectors~\cite{huang2021bevdet, huang2022bevdet4d, li2023bevdepth, li2022bevformer, li2023bevstereo, yang2023bevformer} transform image features into Bird's-Eye-View (BEV) space and directly perform detection on the BEV features, demonstrating competitive performance. This illustrates the substantial advantages of BEV representation in preserving comprehensive scene information, making it more adept for vision-centric autonomous driving perception than isolated image features in perspective space~\cite{park2021pseudo, wang2021fcos3d, wang2022probabilistic, wang2022detr3d}.

\begin{figure}[t]
    \centering
    \subfloat[Baseline]{ \includegraphics[width=0.45\linewidth]{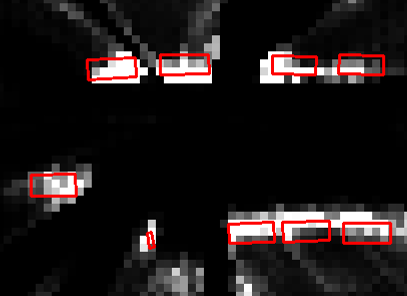}} 
    \subfloat[Larger BEV size]{ \includegraphics[width=0.45\linewidth]{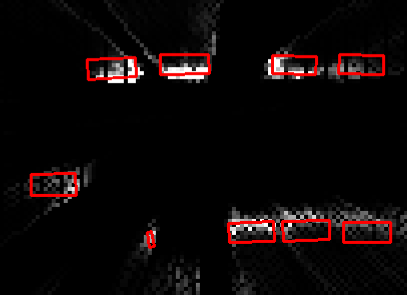}}  \\
    \subfloat[Applying RC-Sampling]{ \includegraphics[width=0.45\linewidth]{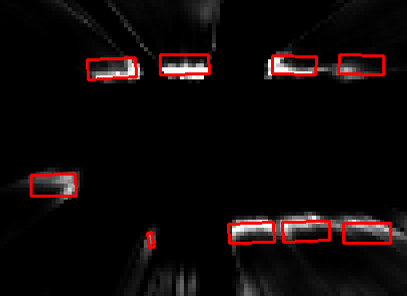}}
    \subfloat[Applying In-Box Label]{ \includegraphics[width=0.45\linewidth]{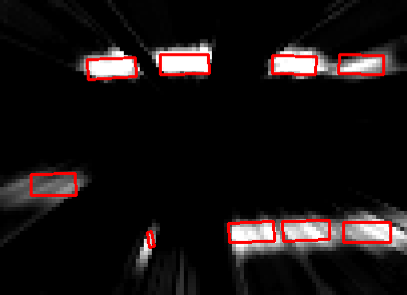}}
    \caption{Comparison between BEV representations. BEVDepth is chosen as the baseline. Larger BEV size, RC-Sampling and In-Box Label are added in turn. The boxes represent the ground truth of the scene and brightness reveals the norm of the features. The background is filtered out to show the difference in the foreground.}
    \label{fig:bevfeat_compare}
\end{figure}

As the cornerstone of BEV-based approaches, the BEV representation embodies both contextual semantic information and depth geometric information. The former is derived from image features, while the latter originates from the correlation between image features and BEV features. Both types of information are indispensable for precise 3D object detection. However, the geometric quality of BEV representation has never received sufficient attention, and the limitation of low-resolution representation always arises. LSS-based methods ~\cite{xie2022m, philion2020lift, huang2021bevdet, huang2022bevdet4d} pool pseudo-points into BEV representation, leaving the positions without pseudo-points to have vacant features. The sparsity will further increase along with the BEV resolution as shown in Fig.~\ref{fig:bevfeat_compare}(b). For Transformer-based methods~\cite{li2022bevformer, yang2023bevformer, jiang2023polarformer, li2023dfa3d} that employ cross-attention to retrieve image features, the elevated BEV resolution leads to a rapid escalation in computational costs. The lack of explicit depth distribution also limits their ability to restore accurate geometric information. Some methods~\cite{harley2023simple, peng2023learning} simply sample the image features to obtain voxel features, which are then squeezed into BEV representation. However, this kind of feature transformation, which we call Voxel-Sampling, requires a large number of sampling operations and produces huge intermediate features, giving it no advantage over LSS-based and Transformer-based methods.

To solve the drawbacks of the existing feature transformation mechanisms, we propose Radial-Cartesian BEV Sampling (RC-Sampling) to generate dense BEV representation with high resolution efficiently. Initially, we create Radial BEV features correlated to each camera view by extending the depth dimension and squeezing the height dimension of image features. We prove that this step can be achieved by simple matrix transposition and multiplication without creating huge intermediate voxel features. Subsequently, bilinear sampling is employed to retrieve the corresponding Radial BEV features for populating the BEV features in Cartesian coordinates. The number of sampling operations between different BEV features is far less than the sampling operations used to create voxel features. RC-Sampling creates the BEV representation of the same quality as Voxel-Sampling while reducing more than 90\% time cost and memory cost according to our experiments. RC-Sampling is also faster than the most efficient LSS approach like BEVPoolv2~\cite{huang2022bevpoolv2}, which relies on custom operator acceleration, and thoroughly solves the problem of the feature vacancy as shown in Fig.~\ref{fig:bevfeat_compare}(c).

Truthfully representing the real spatial distribution of the objects is as important as increasing the BEV resolution. Some methods supervise predicted depth scores by utilizing the depth values of LiDAR points as depth labels~\cite{reading2021categorical, li2023bevdepth, li2023bevstereo, wang2022sts, zhang2023sa, jiang2025fsd}. However, the LiDAR labels only record the depth of object surfaces that face the ego car, failing to represent the actual geometric structure of objects in real-world space. We propose In-Box Label to offer more competent supervision. We first check whether the generated pseudo-points are within the GT boxes and obtain binary labels. These labels, called Vanilla In-Box Label, can effectively incentivize the network to assign high depth scores to where the objects are actually located. Nonetheless, they may lead to feature confusion caused by object occlusion or wrongly boxed background pseudo-points. We ameliorate those issues to enhance its accuracy in reflecting the geometric structure of the scene. In conjunction with the utilization of In-Box Labels, Centroid-Aware Inner Loss (CAI Loss) is also proposed to capture the fine-grained inner geometric structure of objects. After applying the In-Box Label, the authentic geometric structures of objects are clearly presented as shown in Fig.~\ref{fig:bevfeat_compare}(d), and more precise detection is facilitated. It is noteworthy that both In-Box labels and CAI Loss do not introduce extra parameters.

We integrate the aforementioned modules into a novel multi-view 3D object detector, dubbed GeoBEV, and carry out extensive experiments on the nuScenes dataset. The major contributions of this paper can be summarized as:
\begin{itemize}
    \item We propose Radial-Cartesian BEV Sampling to conveniently acquire Cartesian BEV features by bilinearly sampling Radial BEV features, which enables the efficient generation of high-resolution dense BEV representation, facilitating the recovery of fine-grained geometric details within the scene.
    \item We design the novel In-Box Label, cooperating with Centroid-Aware Inner Loss, to supervise the depth scores, which better reflects the actual geometric structure of the object than the LiDAR label and inject authentic geometric information into the BEV representation.
    \item Extensive experiments are conducted on the nuScenes Dataset, and GeoBEV reaches newly state-of-the-art results of 66.2\% NDS for multi-view 3D object detection, highlighting its effectiveness.
\end{itemize}

\section{Related Work}
\subsection{Depth Prediction Based BEV Representation}

Due to the limitation of the camera in capturing the depth required for 3D object detection, predicting the depth distribution of image elements becomes a natural choice. Early methods like OFT~\cite{roddick2018orthographic} assume that the depth distribution of image elements is uniform and all voxels along the ray share the same features. Lately, LSS~\cite{philion2020lift} enables adaptive depth prediction and weights image features to generate pseudo-points at corresponding depth values, which are pooled into BEV features. BEVDet~\cite{huang2021bevdet} employs LSS to construct the detection framework and applies data augmentation in the BEV space. BEVDet4D~\cite{huang2022bevdet4d} merges the BEV features from past frames to help predict the objects' velocity.

In order to obtain more accurate depth information, CaDDN~\cite{reading2021categorical} projects the LiDAR points onto the image to supervise the predicted depth distribution. BEVDepth~\cite{li2023bevdepth} considers the camera's internal and external parameters and further optimizes the depth distribution after the supervision. BEVStereo~\cite{li2023bevstereo} introduces multi-view stereo to obtain more reliable depth distributions and performs some optimizations to minimize memory usage. TiG-BEV~\cite{huang2022tig} sets key points to learn the local depth structure of the scene. SA-BEV~\cite{zhang2023sa} segments the images to get the foreground-only BEV features and improves depth distribution via multi-task learning. BEV-IO~\cite{zhang2023bev} adopts instance occupancy prediction modules as a complement to depth prediction. FB-BEV~\cite{li2023fb} adds a backward process to fill a part of vacant features. BEVNext~\cite{li2024bevnext} adopts CRF to modulate the estimated depth. However, these attempts fail to record the actual object structure due to the limitation of LiDAR points. 

\begin{figure*}[t]
    \centering
    \includegraphics[width=0.75\linewidth]{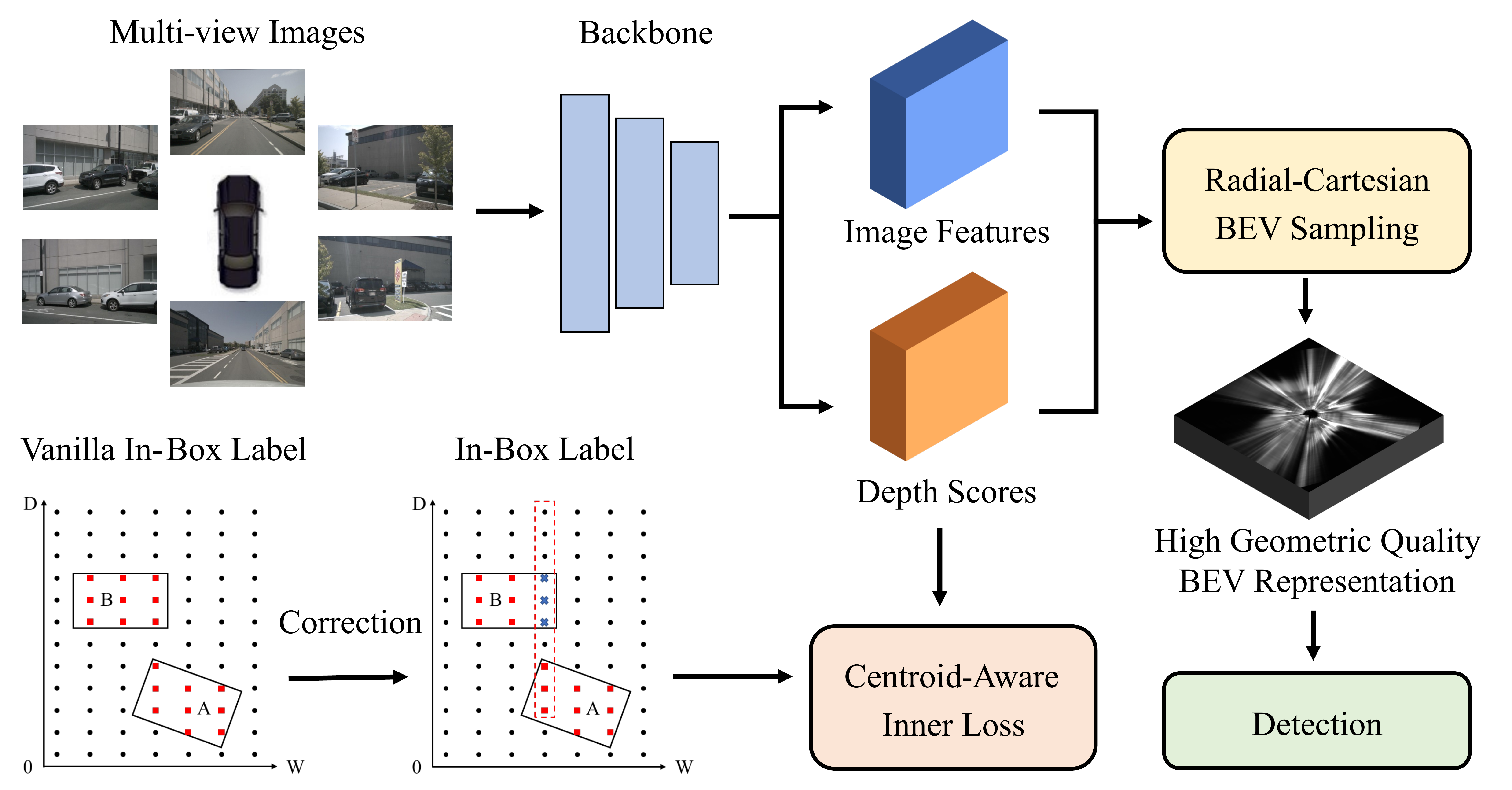}
    \caption{Overall architecture of GeoBEV. The multi-view images are processed into the image features and depth scores. The depth scores are supervised by the In-Box Label that records authentic geometric structures of objects through the Centroid-Aware Inner Loss. Radial-Cartesian BEV Sampling then efficiently generates dense BEV representation with high resolution.}
    \label{fig:geobev}
\end{figure*}

\subsection{Transformer Based BEV Representation}
With the attention mechanism, Transformer-based detectors can adaptively retrieve image features to obtain dense BEV representation. BEVFormer~\cite{li2022bevformer} uses deformable attention to transform image features into BEV space and fuse the past BEV representation. 
BEVFormerV2~\cite{yang2023bevformer} introduces a detection head in perspective view to make the image features more suitable for 3D detection. PolarFormer~\cite{jiang2023polarformer} generates the BEV representation in polar coordinates, which are more competent for ego car perception. DFA3D~\cite{li2023dfa3d} utilizes the explicit depth distribution in cross attention and simplifies the 3D Transformer into the 2D Transformer equivalently.

Several Transformer-based detectors regard the objects as queries to save the large amount of computation required to generate the explicit BEV representation. DETR3D~\cite{wang2022detr3d} follows DETR series detectors~\cite{carion2020end, zhu2020deformable} and interacts object queries with multi-view image features. PETR~\cite{liu2022petr} embeds 3D position into the image features, supplementing spatial information to the object queries. PETRv2~\cite{liu2023petrv2} extends PETR for temporal modeling and adds map queries for other perception tasks. StreamPETR~\cite{wang2023exploring} propagates long-term historical information. Sparse4D~\cite{lin2022sparse4d} assigns multiple 4D key points to aggregate multi-view/scale/timestamp image features. Sparse4Dv2~\cite{lin2023sparse4d} uses the recurrent method to transmit the temporal information. 
RayDN~\cite{liu2024ray} constructs positive and negative examples along the camera rays to learn depth-aware features.
Nevertheless, the omission of explicit BEV representation causes geometric information loss, limiting their precision upper bound. 

\section{Method}
\subsection{Overall Architecture}

The overall architecture of our proposed GeoBEV is shown in Fig.~\ref{fig:geobev}. Firstly, the multi-view images are processed by the image backbone and DepthNet to provide the image features and depth scores. Then the In-Box Label is created and utilized to supervise the depth scores to restore the actual distribution of the objects in BEV space. Centroid-Aware Inner Loss is adopted to let the model learn the inner structure of the objects. Finally, Radial-Cartesian BEV Sampling generates dense BEV representation with high resolution, outperforming current feature transformation approaches in both efficiency and effectiveness.

\begin{figure}[t]
    \centering
    \includegraphics[width=0.9\linewidth]{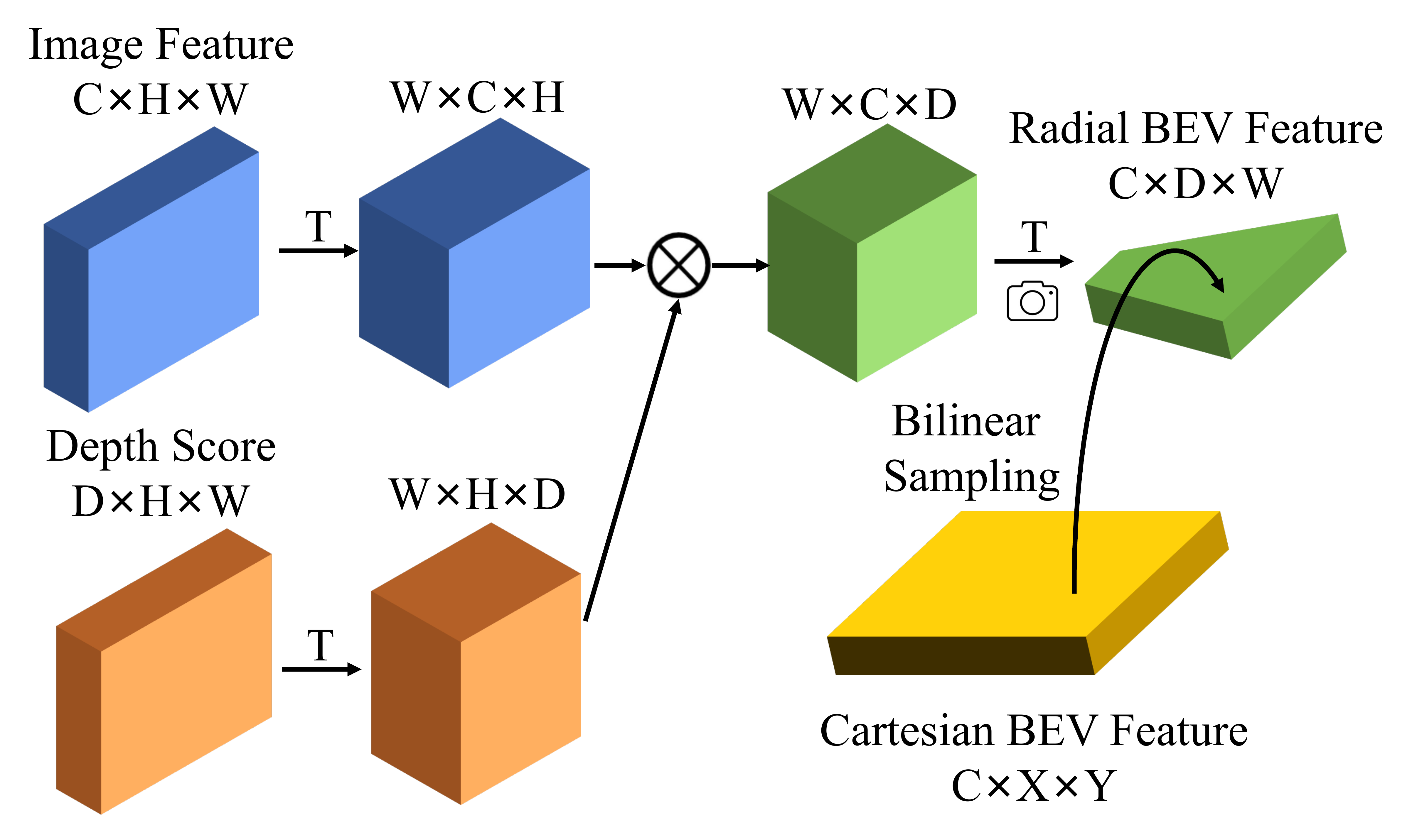}
    \caption{The illustration of Radial-Cartesian BEV Sampling. After high-dimensional matrix multiplication between the transposed image features and depth scores, the $H$ dimension is squeezed to obtain Radial BEV features.}
    \label{fig:rcsampling}
\end{figure}

\subsection{Radial-Cartesian BEV Sampling}
The drawbacks of current methods limit the resolution of BEV representation, failing to restore the fine-grained geometric information of the scene. For LSS-based methods, the density imbalance of pseudo-points leads to feature vacancy in BEV representation, which will deteriorate further as the BEV resolution increases. FB-BEV~\cite{li2023fb} applies backward projection to fill these vacant features but relies on imprecise RoI predicted from the sparse BEV features. The cross-attention between the image and BEV space in Transformer-based methods guarantees the density of BEV features, but the computational cost increases rapidly along with the BEV resolution. Some methods ~\cite{peng2023learning, harley2023simple} sample image features following strict projection relations to obtain voxel features, which are summarized along the height into dense BEV features. However, the large number of sampling operations and the huge intermediate features lessen their practicability.

Here, we propose Radial-Cartesian BEV Sampling (RC-Sampling) to generate dense BEV features with high resolution conveniently. Firstly, the Radial BEV features corresponding to each image view are created by extending the depth dimension and summarizing the height dimension of the image features. They are so-called because their elements appear radially distributed after being projected into the 3D space as shown in Fig~\ref{fig:rcsampling}.
We denote image features as $\bold{I}\in\mathbb{R}^{\scriptscriptstyle C\times H\times W}$ and depth scores as $\bold{D}\in\mathbb{R}^{\scriptscriptstyle D\times H\times W}$, where $C, D, H, W$ represent the channel, depth, height and width dimension respectively. The elements in Radial BEV features $\bold{B}^R\in\mathbb{R}^{\scriptscriptstyle C\times D\times W}$ can be represented by:
\begin{equation}
    \gamma_{cdw}=\sum_{h\in H} \alpha_{chw}\beta_{dhw},
    \label{eq:1}
\end{equation}
where $\alpha, \beta, \gamma$ are the elements of $\bold{I}, \bold{D}, \bold{B}^R$, and $c, d, h, w$ are the indexes of $C, D, H, W$. The general approach is to create 4D frustum features $\bold{F}\in\mathbb{R}^{\scriptscriptstyle C\times D\times H\times W}$ and summarize along the $H$ dimension. By contrast, RC-Sampling implements this process more efficiently.

Omitting the common dimension $W$, the Equation~\ref{eq:1} can be simplified as:
\begin{equation}
    \gamma_{cd}=\sum_{h\in H} \alpha_{ch}\beta_{dh} \Rightarrow 
    \bold{B}^R_w = \bold{I}^{}_w\bold{D}^\top_w,
\end{equation}
where $\bold{B}^R_w\in \mathbb{R}^{\scriptscriptstyle C\times D}, \bold{I}_w\in \mathbb{R}^{\scriptscriptstyle C\times H}, \bold{D}_w\in \mathbb{R}^{\scriptscriptstyle D\times H}$ are the slices of $\bold{I}, \bold{D}, \bold{B}^R$ at $w$. After considering dimension $W$, $\bold{B}^R$ can be directly created by:
\begin{equation}
    \bold{B}^R = [(\bold{I}\rightarrow \mathbb{R}^{\scriptscriptstyle W\times C\times H})\otimes(\bold{D}\rightarrow \mathbb{R}^{\scriptscriptstyle W\times H\times D})]\rightarrow \mathbb{R}^{\scriptscriptstyle C\times D\times W},
\end{equation}
where $\rightarrow$ and $\otimes$ denote the transposition and multiplication of the high-dimension matrix. As shown in Fig.~\ref{fig:rcsampling}, additional computation and memory required by frustum features $\bold{F}$ are saved for equally creating $\bold{B}^R$.

The $\bold{B}^R$ needs to be transformed into Cartesian coordinates for subsequent detection. We pre-define the coordinates of Cartesian BEV features $\bold{B}^C\in\mathbb{R}^{\scriptscriptstyle C\times X\times Y}$, where $X, Y$ denote the required BEV resolution, and project them on the $\bold{B}^R$. The bilinear sampling is utilized to retrieve the corresponding features, which can be represented by :
\begin{equation}
    \bold{B}^C(x,y)={\rm BilinearSample}(\bold{B}^R, {\rm Project}(x,y)),
\end{equation}
where ${\rm Project}(x,y)$ denotes the coordinates of the projected point $(x,y)$ on $\bold{B}^R$. 
Bilinearly sampling $\bold{B}^R$ instead of pooling the sparse pseudo-points guarantees that each position in $\bold{B}^C$ has valid features. It also saves more than 90\% time cost and memory cost required by Voxel-Sampling while creating BEV representation with equal geometric quality. The efficiency and quality advantages can be maintained when the BEV resolution is increased. Along with larger $\bold{B}^C$, $\bold{I}$ and $\bold{D}$ are also enlarged by lightweight convolution to provide fine-grained information of the scene.

Compared with other feature transformation methods, RC-Sampling does not require the generation of memory-expensive 3D intermediate features, the utilization of deployment-unfriendly custom operators or the computation-expensive cross-attention mechanism, highlighting its usability. Experiment results illustrate that RC-Sampling outperforms the state-of-the-art feature transformation methods, such as BEVPoolv2 and DFA3D, on both precision and efficiency.

\begin{figure}[t]
    \centering
    \subfloat[Occlusion Between Objects]{ \includegraphics[width=0.48\linewidth]{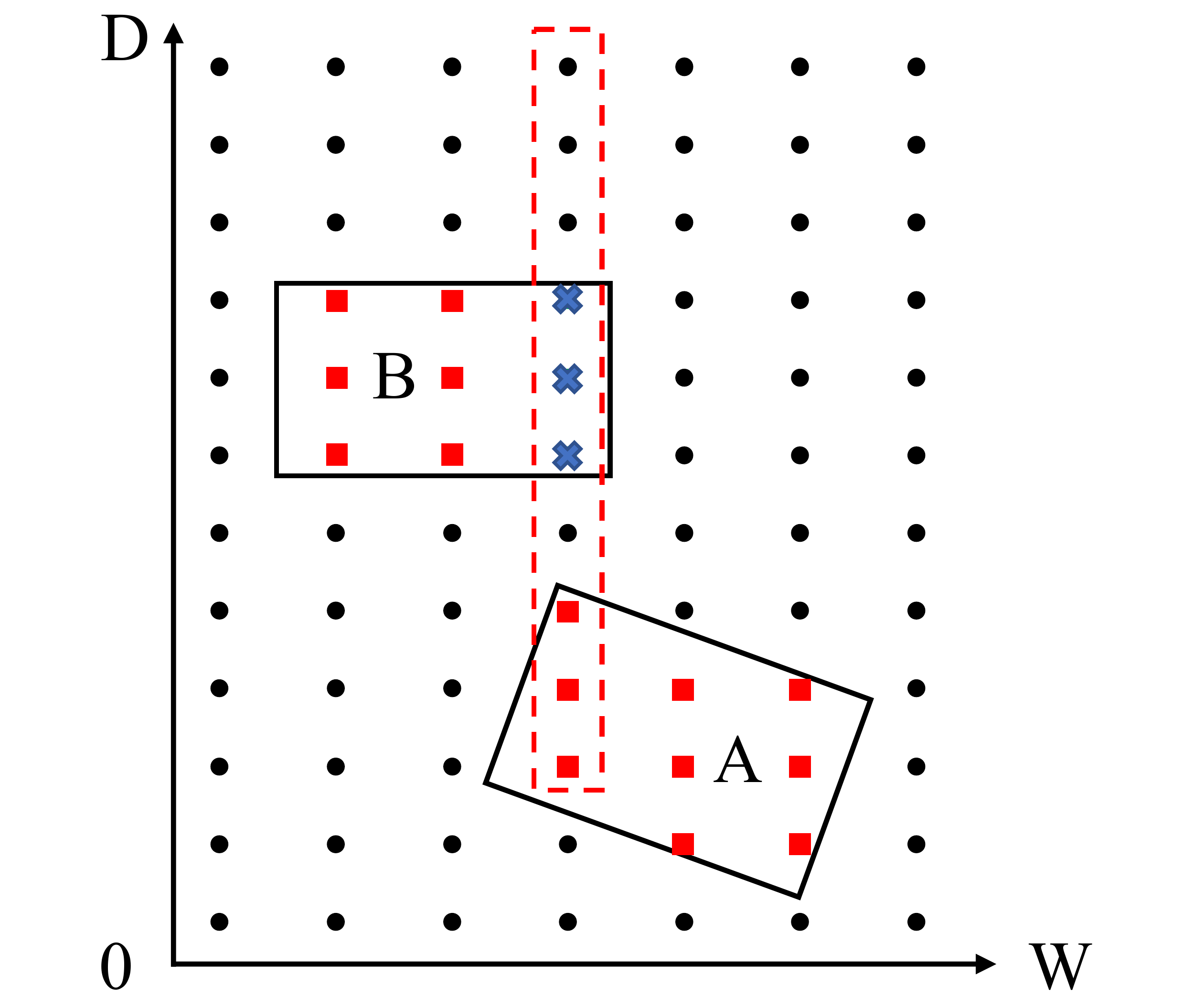} }
    \subfloat[Background in Box]{ \includegraphics[width=0.48\linewidth]{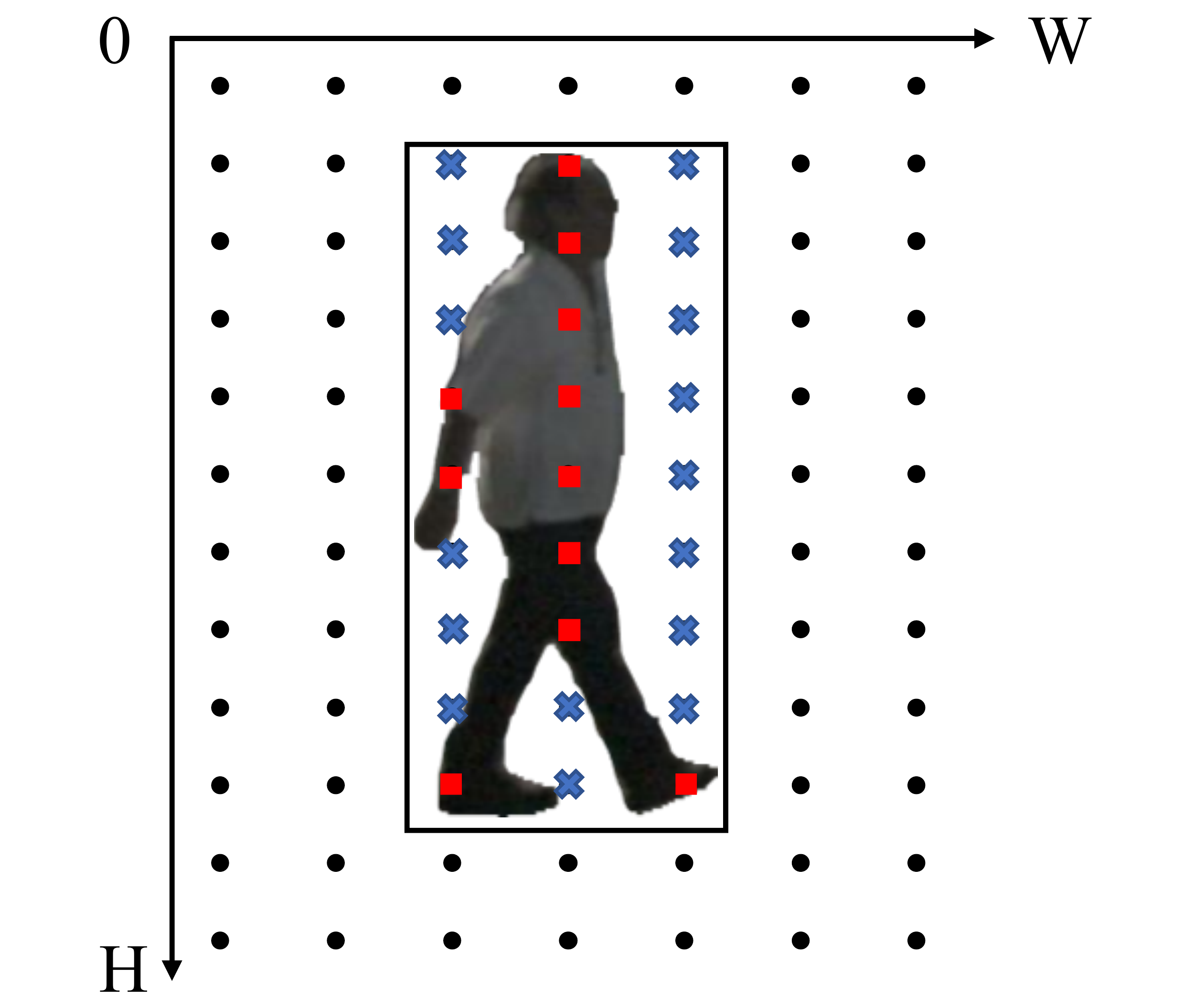} }
    \\
    \subfloat[Behind Background Surface]{ \includegraphics[width=0.48\linewidth]{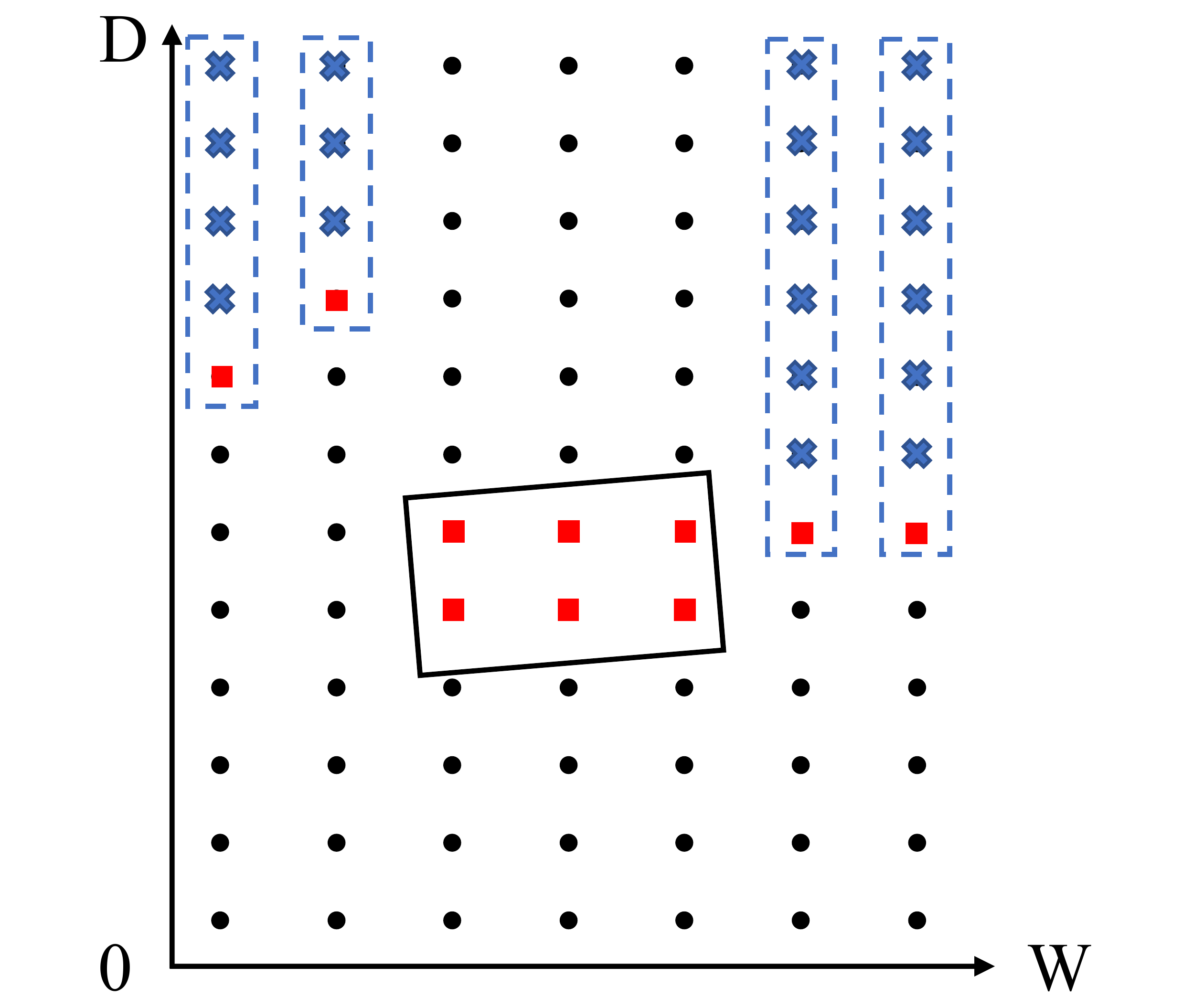}}
    \subfloat[Centroid-Aware Inner Loss]{ \includegraphics[width=0.48\linewidth]{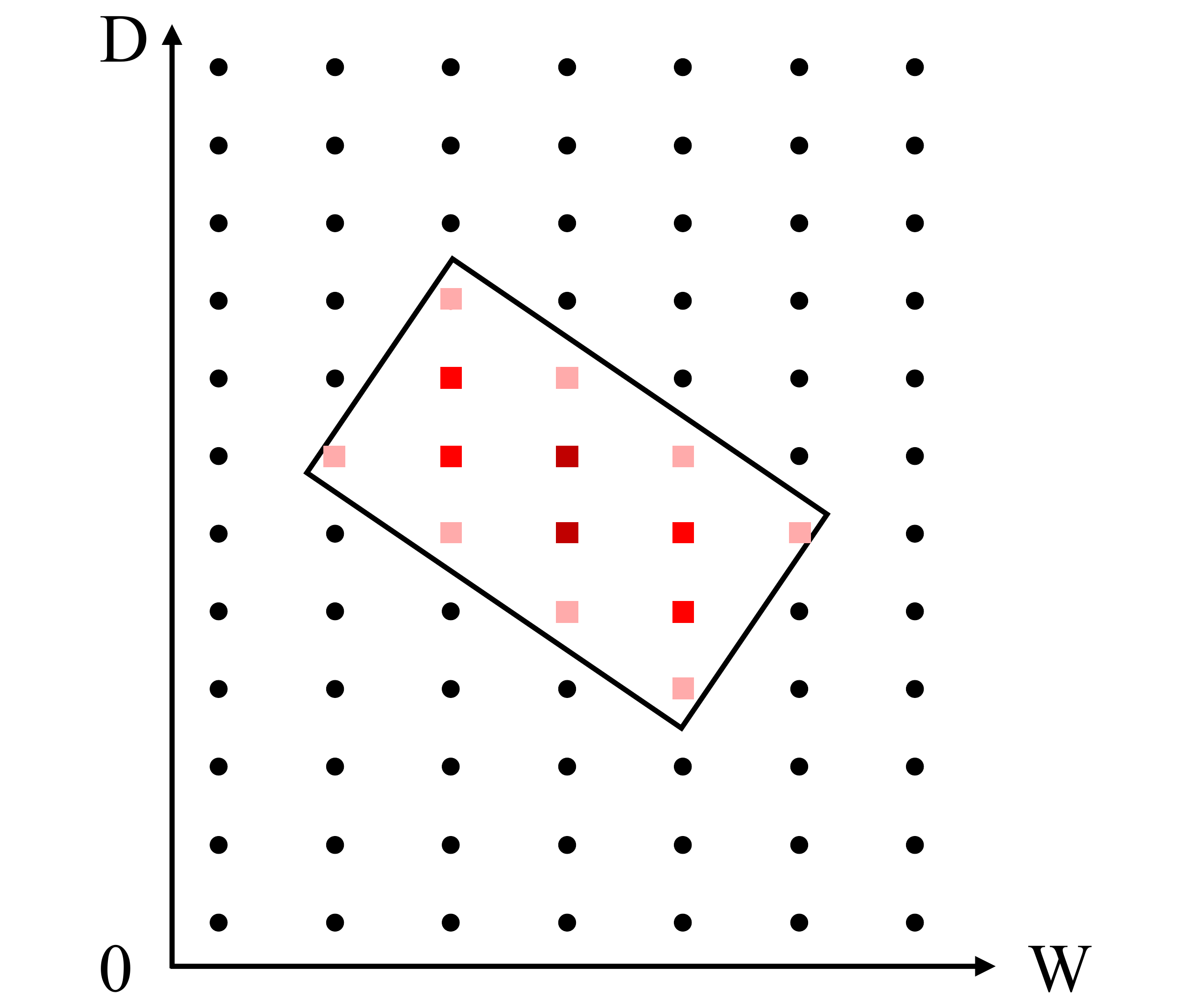} }
    \caption{Illustration of the associated design of In-Box Label. $H, W, D$ represent the height, width and depth dimensions. The boxes are the GT boxes. The red squares and black dots denote the positive and negative points of the In-Box Label. The blue crosses are the points that are not supervised. The deeper color in (d) means higher loss weight.}
    \label{fig:inbox_label}
\end{figure}
\subsection{In-Box Label}

LiDAR points have been used to supervise the depth score of each pseudo-points, effectively attaching the geometric information to the BEV representation. However, the LiDAR label only records the depth of the object surfaces facing the ego car, instead of the actual geometric structure of the objects. The lack of objects' complete geometric information hinders the subsequent BEV encoder and detection head from precisely recognizing their size and orientation.
To overcome the drawbacks of the LiDAR label, we propose the In-Box Label that can be easily obtained from the 3D coordinate of pseudo-points and the GT boxes.

Denote the 3D coordinate of a pseudo-point generated from image features as $p\in\mathbb{R}^3$ and the space within a GT box as $B$, the Vanilla In-Box Label can be formulated as:
\begin{equation}
    L_{inbox} = 
        \begin{cases}
            1,& p\in\bigcup\limits_{i=1}^NB_i \\
            0,& p\notin\bigcup\limits_{i=1}^NB_i
        \end{cases}
\end{equation}
where $N$ is the total number of GT boxes. It means if $p$ is within any GT boxes, it is regarded as positive. Such depth labels encourage the model to describe the actual geometric structure of objects well and fill the GT boxes with valid features in BEV space as shown in Fig.~\ref{fig:bevfeat_compare}(d).

However, Vanilla In-Box Label may cause mismatches between image features and BEV representation of objects and several corrections are needed. For instance, since Object A in Fig.~\ref{fig:inbox_label}(a) has a smaller depth than Object B, the image records the information of Object A in the occluded area (represented by the red dotted box). If the blue crosses are treated as positive, the network will give a high depth score there and mix Object B with A, which is harmful to perception. We choose not to supervise the pseudo-points within the occluded region and let the network learn to give a proper depth score by itself. A similar solution is adopted when objects have irregular shapes, as shown in Fig.~\ref{fig:inbox_label}(b). Not all pseudo-points within the GT box record the information object and they should also be ignored during training. We use the HTC~\cite{chen2019hybrid} pre-trained on nuImages~\cite{caesar2020nuscenes} to provide the mask of objects and filter out the background pseudo-points while calculating loss.

As for the background regions where no GT boxes are available, the LiDAR label is still employed to make the network learn the whole depth distribution of the scene and locate objects more precisely. Since the LiDAR label reflects the depth of the surfaces while In-Box Label records the actual spatial distribution, we modify the LiDAR label to resolve the optimization divergence between foreground and background. As shown in Fig.~\ref{fig:inbox_label}(c), we also ignore the pseudo-points behind the background surface (represented by blue dotted boxes), which are used to be negative. It lets the network adaptively predict how ``thick'' is the ground and the surrounding buildings and also balances the number of positive and negative.

\begin{table*}[t]
    \begin{center}  
        \small
        \setlength{\tabcolsep}{1.4mm}{
        \begin{tabular}{l|c|c|c|cc|ccccc}
            \hline
            Method & Backbone & Image Size & Frames & mAP$\uparrow$ & NDS$\uparrow$ & mATE$\downarrow$ & mASE$\downarrow$ & mAOE$\downarrow$ & mAVE$\downarrow$ & mAAE$\downarrow$\\    
            \hline
            BEVDet~\cite{huang2021bevdet} & ResNet50 & 256$\times$704 & 1 & 0.298 & 0.379 & 0.725 & 0.279 & 0.589 & 0.860 & 0.245\\
            PETRv2 \cite{liu2023petrv2} & ResNet50 & 256$\times$704 & 2 & 0.349 & 0.456 & 0.700 & 0.275 & 0.580 & 0.437 & \textbf{0.187}\\
            BEVDepth \cite{li2023bevdepth} & ResNet50 & 256$\times$704 & 2 & 0.351 & 0.475 & 0.639 & 0.267 & 0.479 & 0.428 & 0.198\\
            BEVStereo \cite{li2023bevstereo} & ResNet50 & 256$\times$704 & 2 & 0.372 & 0.500 & 0.598 & 0.270 & 0.438 & 0.367 & 0.190\\
            SA-BEV~\cite{zhang2023sa} & ResNet50 & 256$\times$704 & 2 & 0.387 & 0.512 & 0.613 & 0.266 & \textbf{0.352} & 0.382 & 0.199 \\
            BEVFormerv2 \cite{yang2023bevformer} & ResNet50 & -  & - & 0.423& 0.529  &0.618  &0.273  &0.413  &0.333  &0.188\\
            SOLOFusion \cite{park2022time} & ResNet50 & 256$\times$704 & 17 & 0.427 & 0.534 & 0.567 & 0.274 & 0.511 & 0.252 & 0.181\\
            StreamPETR$^\star$~\cite{wang2023exploring} & ResNet50 & 256$\times$704 & 8 & 0.450 & 0.550 & 0.613 & 0.267 & 0.413 & 0.265 & 0.196\\
            BEVNext$^\star$~\cite{li2024bevnext} & ResNet50 & 256$\times$704 & 8 & 0.456 & 0.560 & 0.530 & 0.264 & 0.424 & 0.252 & 0.206 \\
            RayDN$^\star$~\cite{liu2024ray} & ResNet50 & 256$\times$704 & 8 & 0.469 & 0.563 & 0.579 & 0.264 & 0.433 & 0.256 & 0.187 \\
            \hline
            GeoBEV & ResNet50 & 256$\times$704 & 2 & 0.415 & 0.535 & 0.533 & 0.265 & 0.419 & 0.298 & 0.214\\
            GeoBEV$^\star$ & ResNet50 & 256$\times$704 & 8 & \textbf{0.479} & \textbf{0.575} & \textbf{0.496} & \textbf{0.261} & 0.438 & \textbf{0.236} & 0.216\\
            \hline 
            PETRv2 \cite{liu2023petrv2} & ResNet101& 900$\times$ 1600 & 2 & 0.421 & 0.524 & 0.681 & 0.267 & 0.357 & 0.377 & \textbf{0.186} \\  
            BEVDepth~\cite{li2023bevdepth} & ResNet101 & 512$\times$1408 & 2 & 0.412 & 0.535 & 0.565 & 0.266 & 0.358 & 0.331 & 0.190 \\
            SOLOFusion~\cite{park2022time} & ResNet101 & 512$\times$1408 & 17 & 0.483 & 0.582 & 0.503 & 0.264 & 0.381 & 0.246 & 0.207 \\
            StreamPETR$^\star$~\cite{wang2023exploring} & ResNet101 & 512$\times$1408 & 8 & 0.504 & 0.592 & 0.569 & 0.262 & \textbf{0.315} & 0.257 & 0.199 \\
            BEVNext$^\star$~\cite{li2024bevnext} & ResNet101 & 512$\times$1408 & 8 & 0.500 & 0.597 & 0.487 & 0.260 & 0.343 & 0.245 & 0.197 \\
            RayDN$^\star$~\cite{liu2024ray} & ResNet101 & 512$\times$1408 & 8 & 0.518 & 0.604 & 0.541 & 0.260 & \textbf{0.315} & 0.236 & 0.200 \\
            \hline
            GeoBEV & ResNet101 & 512$\times$1408 & 2 & 0.479 & 0.582 & 0.498 & \textbf{0.254} & 0.335 & 0.285 & 0.204\\
            GeoBEV$^\star$ & ResNet101 & 512$\times$1408 & 8 & \textbf{0.526} & \textbf{0.615} & \textbf{0.458} & \textbf{0.254} & 0.318 & \textbf{0.238} & 0.207\\
            \hline 
        \end{tabular}}
        \caption{Comparison with previous state-of-the-art multi-view 3D detectors on the nuScenes \textit{val} set. $^\star$ Benefited from the perspective-view pre-training.}
        \label{tab:val}
    \end{center}
\end{table*}

\subsection{Centroid-Aware Inner Loss}

Adapted to the characteristics of In-Box Label, we propose Centroid-Aware Inner Loss (CAI Loss) to replace the Cross-Entropy depth loss. It encourages the model to learn the inner structure of the objects and further refine the geometric information of BEV representation.

Softmax and Cross-Entropy Loss are formally chosen as the activation function and the depth loss to match the one-hot LiDAR label. Softmax centralizes the depth score on the depth values of the object surfaces and Cross-Entropy Loss treats discrete depth values as different classes.
When In-Box Label is utilized, the network should give all the pseudo-points within the GT boxes high depth scores. We choose Sigmoid to independently normalize the depth scores of each pseudo-point within $[0,1]$. 
Besides, the multiple classification of discrete depth values turns into the binary classification of whether pseudo-points are in boxes, which results in far more negative than positive. As a result, Focal Loss~\cite{lin2017focal} is adopted to balance the losses of different classes. 

To learn the inner structure of objects, we vary the loss weights of positive pseudo-points according to their relative position in the GT boxes. Inspired by ~\cite{zhang2022not}, Centroid-Aware Inner Weight is defined as:
\begin{equation}
    W_{\scriptscriptstyle CAI}=\sqrt[3]{\frac{\min(f,b)}{\max(f,b)}\times\frac{\min(l,r)}{\max(l,r)}\times\frac{\min(u,d)}{\max(u,d)}},
\end{equation}
where $f,b,l,r,u,d$ represent the distance of a pseudo-point to the front, back, left, right, up and down surfaces of the GT box. Only the weight of positive pseudo-points needs to be calculated and the pseudo-point closer to the centroid of an object will have a higher weight as shown in Fig.~\ref{fig:inbox_label}(d). The weights are directly multiplied over the focal loss of positive pseudo-points and the CAI Loss is calculated by: 
\begin{equation}
    \mathcal{L}_{\scriptscriptstyle CAI}(p,y)=
        \begin{cases}
            -(1-\alpha)p^{\gamma}\log(1-p),& y=0 \\
            -W_{\scriptscriptstyle CAI}\alpha(1-p)^{\gamma}\log(p),& y=1
        \end{cases},
\end{equation}
where $y$ and $p$ are the label and the activated depth score, $\alpha$ and $\gamma$ are the parameters of Focal Loss.
CAI loss will let the pseudo-points near the object centroids have higher depth scores than the ones near the GT box surfaces, thus expressing the inner geometric information of objects.

\section{Experiments}

\subsection{Dataset and Metrics} 
We evaluate our proposed method on the nuScenes~\cite{caesar2020nuscenes} dataset, a commonly used autonomous driving benchmark. It contains 1000 scenarios collected from the real world, each lasting for around 20 seconds. The key samples are annotated at 2Hz and each sample is provided with the data collected from six cameras, one LiDAR and five radars. The 1000 scenarios are split into training set (750 scenarios), validation (150 scenarios) and test set (150 scenarios). The main metric of the nuScenes dataset for 3D object detection is the nuScenes Detection Score (NDS). Except for the commonly used mean average precision (mAP), NDS is also related to five metrics that only take true positive objects into account, including mean Average Translation Error (mATE), mean Average Scale Error (mASE), mean Average Orientation Error (mAOE), mean Average Velocity Error (mAVE), mean Average Attribute Error (mAAE). 

\begin{table*}
    \begin{center}  
        \small
        \setlength{\tabcolsep}{1.5mm}{
        \begin{tabular}{l|c|c|c|cc|ccccc}
            \hline
            Method & Backbone & Image Size & Frames & mAP$\uparrow$ & NDS$\uparrow$ & mATE$\downarrow$ & mASE$\downarrow$ & mAOE$\downarrow$ & mAVE$\downarrow$ & mAAE$\downarrow$\\                        
            \hline
            BEVDet~\cite{huang2021bevdet} & Swin-B & 900$\times$1600 & 1 & 0.424 & 0.488 & 0.524 & 0.242 & 0.373 & 0.950 & 0.148\\
            BEVFormer~\cite{li2022bevformer} & VoVNet-99 & 900$\times$1600 & 4 & 0.481 & 0.569 & 0.582 & 0.256 & 0.375 & 0.378 & 0.126\\
            PolarFormer~\cite{jiang2023polarformer}& VoVNet-99 & 900$\times$1600 & 2 &0.493&0.572&0.556&0.256&0.364&0.440&0.127\\
            PETRv2~\cite{liu2023petrv2} & VoVNet-99 & 640$\times$1600 & 2 & 0.490 & 0.582 & 0.561 & 0.243 & 0.361 & 0.343 & 0.120\\
            BEVDepth~\cite{li2023bevdepth} & VoVNet-99 & 640$\times$1600 & 2 & 0.503 & 0.600 & 0.445 & 0.245 & 0.378 & 0.320 & 0.126\\
            BEVStereo~\cite{li2023bevstereo} & VoVNet-99 & 640$\times$1600 & 2 & 0.525 & 0.610 & 0.431 & 0.246 & 0.358 & 0.357 & 0.138\\
            SA-BEV~\cite{zhang2023sa} & VoVNet-99 & 640$\times$1600 & 2 & 0.533 & 0.624 & 0.430 & 0.241 & 0.338 & 0.282 & 0.139\\
            FB-BEV~\cite{li2023fb} & VoVNet-99 & 640$\times$1600 & 10 & 0.537 & 0.624 & 0.439 & 0.250 & 0.358 & 0.270 & 0.128 \\
            StreamPETR~\cite{wang2023exploring} & VoVNet-99 & 640$\times$1600 & 8 & 0.550 & 0.636 & 0.479 & 0.239 & \textbf{0.317} & 0.241 & 0.119 \\
            BEVNext~\cite{li2024bevnext} & VoVNet-99 & 640$\times$1600 & 8 & 0.557 & 0.642 & 0.409 & 0.241 & 0.352 & 0.233 & 0.129 \\
            OPEN~\cite{hou2025open} & VoVNet-99 & 640$\times$1600 & - & 0.567 & 0.644 & 0.456 & 0.244 & 0.325 & 0.240 & 0.129 \\
            RayDN~\cite{liu2024ray} & VoVNet-99 & 640$\times$1600 & 8 & 0.565 & 0.645 & 0.461 & 0.241 & 0.322 & 0.239 & \textbf{0.114} \\
            \hline
            GeoBEV & VoVNet-99 & 640$\times$1600 & 2 & 0.543 & 0.635 & 0.409 & \textbf{0.234} & \textbf{0.317} & 0.284 & 0.122 \\
            GeoBEV & VoVNet-99 & 640$\times$1600 & 8 & \textbf{0.579} & \textbf{0.662} & \textbf{0.369} & \textbf{0.234} & 0.323 & \textbf{0.229} & 0.120 \\
            \hline
        \end{tabular}}
        \caption{Comparison with previous state-of-the-art multi-view 3D detectors on the nuScenes \textit{test} set.}
        \label{tab:test}
    \end{center}
\end{table*}

\subsection{Implementation Details}
We adopt the BEVDepth~\cite{li2023bevdepth} as the baseline to build GeoBEV and compare it with state-of-the-art methods in the commonly used configurations. For the experiments on the nuScenes validation set, the ResNet50 and ResNet101~\cite{he2016deep} are adopted as the backbone to process the images in 256$\times$704 and 512$\times$1408, respectively. When evaluating on the nuScenes test set, the VoVNet-99~\cite{lee2019energy} pre-trained by DD3D~\cite{park2021pseudo} is adopted as the backbone to process the images cropped to 640$\times$1600. These models are trained for 20 epochs with CBGS strategy~\cite{zhu2019class}. Except for regular data augmentation, the BEV-Paste~\cite{zhang2023sa} is adopted to alleviate overfitting during the long training process. Future frames and test-time augmentation are not adopted. For the ablation study, we use ResNet50 as the image backbone and the models are trained for 24 epochs without the CBGS strategy.

\subsection{Main Results}
We compare GeoBEV with previous state-of-the-art multi-view 3D detectors on the nuScenes val and test set. The experiment results in Tab.~\ref{tab:val} show that GeoBEV achieves the best detection accuracy on nuScenes val set at different configurations. When detecting from images in 256$\times$704 and using ResNet50 as the backbone, GeoBEV outperforms RayDN~\cite{liu2024ray}, the previous state-of-the-art, by 1.0\% mAP and 1.2\% NDS. When increasing image resolution to 512$\times$1408 and using ResNet101 as the backbone, GeoBEV stays ahead of the curve and outperforms RayDN by 0.8\% mAP and 1.1\% NDS.

The experiment results on the nuScenes test set are shown in Tab.~\ref{tab:test}. GeoBEV also gets the best performance of 57.9\% mAP / 66.3\% NDS, surpassing StreamPETR~\cite{wang2023exploring} by 2.9\% mAP / 2.6\% NDS and RayDN by 1.4\% mAP / 1.7\% NDS, respectively. Those persuasive experiment results highlight the effectiveness of GeoBEV. 

\subsection{Ablation Study}
\subsubsection{Component Analysis}
We evaluate the contributions of our proposed components and show the results in Tab.~\ref{tab:component}. It can be found that both RC-Sampling and In-Box Label effectively increase the detection accuracy. When using BEVDepth~\cite{li2023bevdepth} as the baseline, there is an improvement of 2.6\% mAP and 3.3\% NDS after applying RC-Sampling. In-Box Label and CAI Loss also boost the performance by 2.2\% mAP and 2.2\% NDS. After combining the two components, the performance is increased by 4.4\% mAP and 4.4\% NDS in total.
To estimate the versatility of our proposed components, we also choose BEVDet~\cite{huang2021bevdet} and BEVStereo~\cite{li2023bevstereo} as the baselines. After adopting RC-Sampling and In-Box Label, their accuracy is improved by 2.7\% mAP / 4.1\% NDS and 3.4\% mAP / 3.9\% NDS respectively.

\begin{table}
    \centering
    \small
    \setlength{\tabcolsep}{3mm}{
    \begin{tabular}{l|cc|cc}
        \hline
        Baseline & RC-Sampling & In-Box & mAP & NDS\\ 
        \hline
        \multirow{3}{*}{BEVDepth} & & & 0.337 & 0.456\\
        & \checkmark & & 0.363 & 0.489 \\
        & & \checkmark & 0.359 & 0.478 \\
        & \checkmark & \checkmark & 0.381 & 0.500 \\
        \hline
        \multirow{2}{*}{BEVDet} & & & 0.283 & 0.350 \\
         & \checkmark & \checkmark & 0.310 & 0.391 \\
        \hline
        \multirow{2}{*}{BEVStereo} & & & 0.354 & 0.474 \\
         & \checkmark & \checkmark & \textbf{0.388} & \textbf{0.513} \\
        \hline
    \end{tabular}}
    \caption{Ablation study of proposed components. ``RC-Sampling'' denotes Radial-Cartesian BEV Sampling and ``In-Box'' denotes the combination of In-Box Label and Centroid-Aware Inner Loss.}
    \label{tab:component}
\end{table}

\begin{table}
    \centering
    \small
    \begin{tabular}{l|c|c|cc|c}
        \hline
        Method & BEV Size & DS & mAP & NDS & FPS\\ 
        \hline
        \multirow{2}{*}{BEVPoolv2} & 128$\times$128 & 16 & 0.337 & 0.456 & 22.7\\
        & 256$\times$256 & 16 & 0.344 & 0.474 & 16.6\\
        \hline
        \multirow{2}{*}{DFA3D} & 128$\times$128 & 16 & 0.335 & 0.455 & 20.2\\
         & 256$\times$256 & 16 & 0.344 & 0.469 & 11.7\\
        \hline
        \multirow{2}{*}{Voxel-Sampling} & 128$\times$128 & 16 & 0.342 & 0.464 & 20.6\\
        & 256$\times$256 & 16 & 0.354 & 0.484 & 13.8\\
        \hline
        \multirow{4}{*}{RC-Sampling} & 128$\times$128 & 16 & 0.344 & 0.465 & 24.8\\
        & 256$\times$256 & 16 & 0.358 & 0.482 & 17.4 \\
        & 256$\times$256 & 8 & \textbf{0.363} & \textbf{0.489} & 17.0\\
        \hline
    \end{tabular}
    \caption{Ablation study of Radial-Cartesian BEV Sampling. ``DS'' denotes the downsample factor from the images to the depth scores. ``FPS'' is the FPS of the whole detector.}
    \label{tab:rcsampling}
\end{table}

\subsubsection{Radial-Cartesian BEV Sampling}
To show the capacity of RC-Sampling, we compare it with the most efficient LSS-based and Transformer-based feature transformation methods. BEVPoolv2~\cite{huang2022bevpoolv2} and DFA3D~\cite{li2023dfa3d} are chosen as the representatives. The comparison with Voxel-Sampling, the unoptimized version of RC-Sampling, is also implemented. All of the feature transformation methods are incorporated into the same BEVDepth model. From the experiment results in Tab.~\ref{tab:rcsampling}, it can be found the detection accuracy of RC-Sampling outperforms both BEVPoolv2 and DFA3D, indicating the better geometric quality of BEV representation. Besides, RC-Sampling exhibits better real-time performance and achieves the best FPS while generating BEV representation with different resolutions. Voxel-Sampling achieves comparable accuracy as RC-Sampling, but its speed is far behind. More details of the comparison with Voxel-Sampling are shown in the supplementary material.
We also upsample the size of depth scores by lightweight convolution to provide fine-grained geometry information to RC-Sampling, which further increases the performance by 0.5\% mAP and 0.7\% NDS without significantly affecting its efficiency.

\begin{table}
    \centering
    \small
    \begin{tabular}{l|ccc|cc}
        \hline
        Label & Sigmoid & Focal & CAI & mAP & NDS \\ 
        \hline
        LiDAR & & & & 0.337 & 0.456 \\
        \hline
        \multirow{3}{*}{Vanilla In-Box} & & & & 0.345 & 0.464 \\
        & \checkmark & & & 0.347 & 0.466 \\
        & \checkmark & \checkmark & & 0.351 & 0.470 \\
        \hline
        \multirow{2}{*}{In-Box}& \checkmark & \checkmark & &  0.356 & 0.474 \\
        & \checkmark & & \checkmark & \textbf{0.359} & \textbf{0.478} \\
        \hline
    \end{tabular}
    \caption{Ablation study of In-Box Label. ``Sigmoid'' denotes using Sigmoid as the activation function. ``Focal'' denotes using the focal loss while ``CAI'' denotes using the Centroid-Aware Inner Loss.}
    \label{tab:inbox}
\end{table}

\subsubsection{In-Box Label}
We conduct experiments to evaluate different configurations when applying the In-Box Label as in shown Tab.~\ref{tab:inbox}. When simply replacing the LiDAR label with the Vanilla In-Box Label, the performance is increased by 0.8\% mAP and 0.8\% NDS. It is further improved by 0.2\% mAP / 0.2\% NDS and 0.4\% mAP / 0.4\% NDS after using Sigmoid as the activation function and letting the depth scores supervised by Focal Loss. We also compare the performance between Vanilla In-Box Label and the complete In-Box Label, the results show that the In-Box Label is more in line with the real world and has an advantage of 0.5\% mAP and 0.4\% NDS. When replacing Focal Loss with Centroid-Aware Inner Loss, there is another 0.3\% mAP and 0.4\% NDS improvement, which illustrates that inner geometric structure is helpful for the detection.

\section{Conclusion}

In this paper, we propose a novel multi-view 3D object detector, namely GeoBEV, which generates BEV representation that restores authentic geometric information of the scene. Radial-Cartesian BEV Sampling simply does high-dimensional matrix multiplication between transposed image features and depth scores to obtain Radial BEV features, which are then transformed into Cartesian BEV features by bilinear sampling. This approach can rapidly generate high-resolution BEV representation while effectively avoiding the presence of vacant feature values. Based on the physics of the real world, In-Box Label can reflect the actual geometric structure of objects, effectively improving the accuracy of the information carried by BEV representation. Centroid-Aware Inner Loss cooperates with In-Box Label to make full of its advantage and also encourages the network to learn the inner geometry of objects.

We conduct extensive experiments on nuScenes dataset and GeoBEV reaches a new state-of-the-art, highlighting the effectiveness of our proposed components in enhancing the geometric quality of BEV representation. Additional experiments also illustrate that these components can be easily integrated into many existing BEV-based detectors and bring stable improvement in accuracy and real-time performance.

\section{Acknowledgment}
This research was supported by Zhejiang Provincial Natural Science Foundation of China under Grant No. LD24F020016 and LQ23F020024, National Natural Science Foundation of China under Grant No. 62302031, and “Pioneer” and “Leading Goose” R\&D Program of Zhejiang under Grant No. 2024C01020.

\bibliography{main}

\begin{thebibliography}{44}
\providecommand{\natexlab}[1]{#1}

\bibitem[{Caesar et~al.(2020)Caesar, Bankiti, Lang, Vora, Liong, Xu, Krishnan, Pan, Baldan, and Beijbom}]{caesar2020nuscenes}
Caesar, H.; Bankiti, V.; Lang, A.~H.; Vora, S.; Liong, V.~E.; Xu, Q.; Krishnan, A.; Pan, Y.; Baldan, G.; and Beijbom, O. 2020.
\newblock nuscenes: A multimodal dataset for autonomous driving.
\newblock In \emph{Proceedings of the IEEE/CVF conference on computer vision and pattern recognition}, 11621--11631.

\bibitem[{Carion et~al.(2020)Carion, Massa, Synnaeve, Usunier, Kirillov, and Zagoruyko}]{carion2020end}
Carion, N.; Massa, F.; Synnaeve, G.; Usunier, N.; Kirillov, A.; and Zagoruyko, S. 2020.
\newblock End-to-end object detection with transformers.
\newblock In \emph{European conference on computer vision}, 213--229. Springer.

\bibitem[{Chen et~al.(2019)Chen, Pang, Wang, Xiong, Li, Sun, Feng, Liu, Shi, Ouyang et~al.}]{chen2019hybrid}
Chen, K.; Pang, J.; Wang, J.; Xiong, Y.; Li, X.; Sun, S.; Feng, W.; Liu, Z.; Shi, J.; Ouyang, W.; et~al. 2019.
\newblock Hybrid task cascade for instance segmentation.
\newblock In \emph{Proceedings of the IEEE/CVF conference on computer vision and pattern recognition}, 4974--4983.

\bibitem[{Harley et~al.(2023)Harley, Fang, Li, Ambrus, and Fragkiadaki}]{harley2023simple}
Harley, A.~W.; Fang, Z.; Li, J.; Ambrus, R.; and Fragkiadaki, K. 2023.
\newblock Simple-bev: What really matters for multi-sensor bev perception?
\newblock In \emph{2023 IEEE International Conference on Robotics and Automation (ICRA)}, 2759--2765. IEEE.

\bibitem[{He et~al.(2016)He, Zhang, Ren, and Sun}]{he2016deep}
He, K.; Zhang, X.; Ren, S.; and Sun, J. 2016.
\newblock Deep residual learning for image recognition.
\newblock In \emph{Proceedings of the IEEE conference on computer vision and pattern recognition}, 770--778.

\bibitem[{Hou et~al.(2025)Hou, Wang, Ye, Liu, Gong, Tan, Ding, Wang, and Bai}]{hou2025open}
Hou, J.; Wang, T.; Ye, X.; Liu, Z.; Gong, S.; Tan, X.; Ding, E.; Wang, J.; and Bai, X. 2025.
\newblock Open: Object-wise position embedding for multi-view 3d object detection.
\newblock In \emph{European Conference on Computer Vision}, 146--162. Springer.

\bibitem[{Huang and Huang(2022{\natexlab{a}})}]{huang2022bevdet4d}
Huang, J.; and Huang, G. 2022{\natexlab{a}}.
\newblock Bevdet4d: Exploit temporal cues in multi-camera 3d object detection.
\newblock \emph{arXiv preprint arXiv:2203.17054}.

\bibitem[{Huang and Huang(2022{\natexlab{b}})}]{huang2022bevpoolv2}
Huang, J.; and Huang, G. 2022{\natexlab{b}}.
\newblock Bevpoolv2: A cutting-edge implementation of bevdet toward deployment.
\newblock \emph{arXiv preprint arXiv:2211.17111}.

\bibitem[{Huang et~al.(2021)Huang, Huang, Zhu, Ye, and Du}]{huang2021bevdet}
Huang, J.; Huang, G.; Zhu, Z.; Ye, Y.; and Du, D. 2021.
\newblock Bevdet: High-performance multi-camera 3d object detection in bird-eye-view.
\newblock \emph{arXiv preprint arXiv:2112.11790}.

\bibitem[{Huang et~al.(2022)Huang, Liu, Zhang, Zhang, Xu, Wang, and Liu}]{huang2022tig}
Huang, P.; Liu, L.; Zhang, R.; Zhang, S.; Xu, X.; Wang, B.; and Liu, G. 2022.
\newblock Tig-bev: Multi-view bev 3d object detection via target inner-geometry learning.
\newblock \emph{arXiv preprint arXiv:2212.13979}.

\bibitem[{Jiang et~al.(2023)Jiang, Zhang, Miao, Zhu, Gao, Hu, and Jiang}]{jiang2023polarformer}
Jiang, Y.; Zhang, L.; Miao, Z.; Zhu, X.; Gao, J.; Hu, W.; and Jiang, Y.-G. 2023.
\newblock Polarformer: Multi-camera 3d object detection with polar transformer.
\newblock In \emph{Proceedings of the AAAI Conference on Artificial Intelligence}, volume~37, 1042--1050.

\bibitem[{Jiang et~al.(2025)Jiang, Zhang, Zhang, Liu, Hu, Wang, and Wang}]{jiang2025fsd}
Jiang, Z.; Zhang, J.; Zhang, Y.; Liu, Q.; Hu, Z.; Wang, B.; and Wang, Y. 2025.
\newblock FSD-BEV: Foreground Self-Distillation for Multi-view 3D Object Detection.
\newblock In \emph{European Conference on Computer Vision}, 110--126. Springer.

\bibitem[{Lee et~al.(2019)Lee, Hwang, Lee, Bae, and Park}]{lee2019energy}
Lee, Y.; Hwang, J.-w.; Lee, S.; Bae, Y.; and Park, J. 2019.
\newblock An energy and GPU-computation efficient backbone network for real-time object detection.
\newblock In \emph{Proceedings of the IEEE/CVF conference on computer vision and pattern recognition workshops}, 0--0.

\bibitem[{Li et~al.(2023{\natexlab{a}})Li, Zhang, Zeng, Liu, Li, Ren, and Zhang}]{li2023dfa3d}
Li, H.; Zhang, H.; Zeng, Z.; Liu, S.; Li, F.; Ren, T.; and Zhang, L. 2023{\natexlab{a}}.
\newblock DFA3D: 3D Deformable Attention For 2D-to-3D Feature Lifting.
\newblock In \emph{Proceedings of the IEEE/CVF International Conference on Computer Vision}, 6684--6693.

\bibitem[{Li et~al.(2023{\natexlab{b}})Li, Bao, Ge, Yang, Sun, and Li}]{li2023bevstereo}
Li, Y.; Bao, H.; Ge, Z.; Yang, J.; Sun, J.; and Li, Z. 2023{\natexlab{b}}.
\newblock Bevstereo: Enhancing depth estimation in multi-view 3d object detection with temporal stereo.
\newblock In \emph{Proceedings of the AAAI Conference on Artificial Intelligence}, volume~37, 1486--1494.

\bibitem[{Li et~al.(2023{\natexlab{c}})Li, Ge, Yu, Yang, Wang, Shi, Sun, and Li}]{li2023bevdepth}
Li, Y.; Ge, Z.; Yu, G.; Yang, J.; Wang, Z.; Shi, Y.; Sun, J.; and Li, Z. 2023{\natexlab{c}}.
\newblock Bevdepth: Acquisition of reliable depth for multi-view 3d object detection.
\newblock In \emph{Proceedings of the AAAI Conference on Artificial Intelligence}, volume~37, 1477--1485.

\bibitem[{Li et~al.(2024)Li, Lan, Alvarez, and Wu}]{li2024bevnext}
Li, Z.; Lan, S.; Alvarez, J.~M.; and Wu, Z. 2024.
\newblock BEVNeXt: Reviving Dense BEV Frameworks for 3D Object Detection.
\newblock In \emph{Proceedings of the IEEE/CVF Conference on Computer Vision and Pattern Recognition}, 20113--20123.

\bibitem[{Li et~al.(2022)Li, Wang, Li, Xie, Sima, Lu, Qiao, and Dai}]{li2022bevformer}
Li, Z.; Wang, W.; Li, H.; Xie, E.; Sima, C.; Lu, T.; Qiao, Y.; and Dai, J. 2022.
\newblock Bevformer: Learning bird’s-eye-view representation from multi-camera images via spatiotemporal transformers.
\newblock In \emph{European conference on computer vision}, 1--18. Springer.

\bibitem[{Li et~al.(2023{\natexlab{d}})Li, Yu, Wang, Anandkumar, Lu, and Alvarez}]{li2023fb}
Li, Z.; Yu, Z.; Wang, W.; Anandkumar, A.; Lu, T.; and Alvarez, J.~M. 2023{\natexlab{d}}.
\newblock Fb-bev: Bev representation from forward-backward view transformations.
\newblock In \emph{Proceedings of the IEEE/CVF International Conference on Computer Vision}, 6919--6928.

\bibitem[{Lin et~al.(2017)Lin, Goyal, Girshick, He, and Doll{\'a}r}]{lin2017focal}
Lin, T.-Y.; Goyal, P.; Girshick, R.; He, K.; and Doll{\'a}r, P. 2017.
\newblock Focal loss for dense object detection.
\newblock In \emph{Proceedings of the IEEE international conference on computer vision}, 2980--2988.

\bibitem[{Lin et~al.(2022)Lin, Lin, Pei, Huang, and Su}]{lin2022sparse4d}
Lin, X.; Lin, T.; Pei, Z.; Huang, L.; and Su, Z. 2022.
\newblock Sparse4d: Multi-view 3d object detection with sparse spatial-temporal fusion.
\newblock \emph{arXiv preprint arXiv:2211.10581}.

\bibitem[{Lin et~al.(2023)Lin, Lin, Pei, Huang, and Su}]{lin2023sparse4d}
Lin, X.; Lin, T.; Pei, Z.; Huang, L.; and Su, Z. 2023.
\newblock Sparse4D v2: Recurrent Temporal Fusion with Sparse Model.
\newblock \emph{arXiv preprint arXiv:2305.14018}.

\bibitem[{Liu et~al.(2024)Liu, Huang, Zhang, Yao, Zhang, Wan, Ye, and Zhou}]{liu2024ray}
Liu, F.; Huang, T.; Zhang, Q.; Yao, H.; Zhang, C.; Wan, F.; Ye, Q.; and Zhou, Y. 2024.
\newblock Ray denoising: Depth-aware hard negative sampling for multiview 3d object detection.
\newblock \emph{arXiv preprint arXiv:2402.03634}, 10.

\bibitem[{Liu et~al.(2022)Liu, Wang, Zhang, and Sun}]{liu2022petr}
Liu, Y.; Wang, T.; Zhang, X.; and Sun, J. 2022.
\newblock Petr: Position embedding transformation for multi-view 3d object detection.
\newblock In \emph{European Conference on Computer Vision}, 531--548. Springer.

\bibitem[{Liu et~al.(2023)Liu, Yan, Jia, Li, Gao, Wang, and Zhang}]{liu2023petrv2}
Liu, Y.; Yan, J.; Jia, F.; Li, S.; Gao, A.; Wang, T.; and Zhang, X. 2023.
\newblock Petrv2: A unified framework for 3d perception from multi-camera images.
\newblock In \emph{Proceedings of the IEEE/CVF International Conference on Computer Vision}, 3262--3272.

\bibitem[{Loshchilov and Hutter(2017)}]{loshchilov2017decoupled}
Loshchilov, I.; and Hutter, F. 2017.
\newblock Decoupled weight decay regularization.
\newblock \emph{arXiv preprint arXiv:1711.05101}.

\bibitem[{Park et~al.(2021)Park, Ambrus, Guizilini, Li, and Gaidon}]{park2021pseudo}
Park, D.; Ambrus, R.; Guizilini, V.; Li, J.; and Gaidon, A. 2021.
\newblock Is pseudo-lidar needed for monocular 3d object detection?
\newblock In \emph{Proceedings of the IEEE/CVF International Conference on Computer Vision}, 3142--3152.

\bibitem[{Park et~al.(2022)Park, Xu, Yang, Keutzer, Kitani, Tomizuka, and Zhan}]{park2022time}
Park, J.; Xu, C.; Yang, S.; Keutzer, K.; Kitani, K.; Tomizuka, M.; and Zhan, W. 2022.
\newblock Time will tell: New outlooks and a baseline for temporal multi-view 3d object detection.
\newblock \emph{arXiv preprint arXiv:2210.02443}.

\bibitem[{Peng et~al.(2023)Peng, Xu, Cheng, Yang, Wu, Qian, Wang, Wu, and Cai}]{peng2023learning}
Peng, L.; Xu, J.; Cheng, H.; Yang, Z.; Wu, X.; Qian, W.; Wang, W.; Wu, B.; and Cai, D. 2023.
\newblock Learning Occupancy for Monocular 3D Object Detection.
\newblock \emph{arXiv preprint arXiv:2305.15694}.

\bibitem[{Philion and Fidler(2020)}]{philion2020lift}
Philion, J.; and Fidler, S. 2020.
\newblock Lift, splat, shoot: Encoding images from arbitrary camera rigs by implicitly unprojecting to 3d.
\newblock In \emph{Computer Vision--ECCV 2020: 16th European Conference, Glasgow, UK, August 23--28, 2020, Proceedings, Part XIV 16}, 194--210. Springer.

\bibitem[{Reading et~al.(2021)Reading, Harakeh, Chae, and Waslander}]{reading2021categorical}
Reading, C.; Harakeh, A.; Chae, J.; and Waslander, S.~L. 2021.
\newblock Categorical depth distribution network for monocular 3d object detection.
\newblock In \emph{Proceedings of the IEEE/CVF Conference on Computer Vision and Pattern Recognition}, 8555--8564.

\bibitem[{Roddick, Kendall, and Cipolla(2018)}]{roddick2018orthographic}
Roddick, T.; Kendall, A.; and Cipolla, R. 2018.
\newblock Orthographic feature transform for monocular 3d object detection.
\newblock \emph{arXiv preprint arXiv:1811.08188}.

\bibitem[{Wang et~al.(2023)Wang, Liu, Wang, Li, and Zhang}]{wang2023exploring}
Wang, S.; Liu, Y.; Wang, T.; Li, Y.; and Zhang, X. 2023.
\newblock Exploring object-centric temporal modeling for efficient multi-view 3d object detection.
\newblock In \emph{Proceedings of the IEEE/CVF International Conference on Computer Vision}, 3621--3631.

\bibitem[{Wang et~al.(2022{\natexlab{a}})Wang, Xinge, Pang, and Lin}]{wang2022probabilistic}
Wang, T.; Xinge, Z.; Pang, J.; and Lin, D. 2022{\natexlab{a}}.
\newblock Probabilistic and geometric depth: Detecting objects in perspective.
\newblock In \emph{Conference on Robot Learning}, 1475--1485. PMLR.

\bibitem[{Wang et~al.(2021)Wang, Zhu, Pang, and Lin}]{wang2021fcos3d}
Wang, T.; Zhu, X.; Pang, J.; and Lin, D. 2021.
\newblock Fcos3d: Fully convolutional one-stage monocular 3d object detection.
\newblock In \emph{Proceedings of the IEEE/CVF International Conference on Computer Vision}, 913--922.

\bibitem[{Wang et~al.(2022{\natexlab{b}})Wang, Guizilini, Zhang, Wang, Zhao, and Solomon}]{wang2022detr3d}
Wang, Y.; Guizilini, V.~C.; Zhang, T.; Wang, Y.; Zhao, H.; and Solomon, J. 2022{\natexlab{b}}.
\newblock Detr3d: 3d object detection from multi-view images via 3d-to-2d queries.
\newblock In \emph{Conference on Robot Learning}, 180--191. PMLR.

\bibitem[{Wang et~al.(2022{\natexlab{c}})Wang, Min, Ge, Li, Li, Yang, and Huang}]{wang2022sts}
Wang, Z.; Min, C.; Ge, Z.; Li, Y.; Li, Z.; Yang, H.; and Huang, D. 2022{\natexlab{c}}.
\newblock Sts: Surround-view temporal stereo for multi-view 3d detection.
\newblock \emph{arXiv preprint arXiv:2208.10145}.

\bibitem[{Xie et~al.(2022)Xie, Yu, Zhou, Philion, Anandkumar, Fidler, Luo, and Alvarez}]{xie2022m}
Xie, E.; Yu, Z.; Zhou, D.; Philion, J.; Anandkumar, A.; Fidler, S.; Luo, P.; and Alvarez, J.~M. 2022.
\newblock M2BEV: Multi-Camera Joint 3D Detection and Segmentation with Unified Birds-Eye View Representation.
\newblock \emph{arXiv preprint arXiv:2204.05088}.

\bibitem[{Yang et~al.(2023)Yang, Chen, Tian, Tao, Zhu, Zhang, Huang, Li, Qiao, Lu et~al.}]{yang2023bevformer}
Yang, C.; Chen, Y.; Tian, H.; Tao, C.; Zhu, X.; Zhang, Z.; Huang, G.; Li, H.; Qiao, Y.; Lu, L.; et~al. 2023.
\newblock BEVFormer v2: Adapting Modern Image Backbones to Bird's-Eye-View Recognition via Perspective Supervision.
\newblock In \emph{Proceedings of the IEEE/CVF Conference on Computer Vision and Pattern Recognition}, 17830--17839.

\bibitem[{Zhang et~al.(2023{\natexlab{a}})Zhang, Zhang, Liu, and Wang}]{zhang2023sa}
Zhang, J.; Zhang, Y.; Liu, Q.; and Wang, Y. 2023{\natexlab{a}}.
\newblock SA-BEV: Generating Semantic-Aware Bird's-Eye-View Feature for Multi-view 3D Object Detection.
\newblock In \emph{Proceedings of the IEEE/CVF International Conference on Computer Vision}, 3348--3357.

\bibitem[{Zhang et~al.(2022)Zhang, Hu, Xu, Ma, Wan, and Guo}]{zhang2022not}
Zhang, Y.; Hu, Q.; Xu, G.; Ma, Y.; Wan, J.; and Guo, Y. 2022.
\newblock Not all points are equal: Learning highly efficient point-based detectors for 3d lidar point clouds.
\newblock In \emph{Proceedings of the IEEE/CVF Conference on Computer Vision and Pattern Recognition}, 18953--18962.

\bibitem[{Zhang et~al.(2023{\natexlab{b}})Zhang, Wang, Wang, and Lu}]{zhang2023bev}
Zhang, Z.; Wang, L.; Wang, Y.; and Lu, H. 2023{\natexlab{b}}.
\newblock BEV-IO: Enhancing Bird's-Eye-View 3D Detection with Instance Occupancy.
\newblock \emph{arXiv preprint arXiv:2305.16829}.

\bibitem[{Zhu et~al.(2019)Zhu, Jiang, Zhou, Li, and Yu}]{zhu2019class}
Zhu, B.; Jiang, Z.; Zhou, X.; Li, Z.; and Yu, G. 2019.
\newblock Class-balanced grouping and sampling for point cloud 3d object detection.
\newblock \emph{arXiv preprint arXiv:1908.09492}.

\bibitem[{Zhu et~al.(2020)Zhu, Su, Lu, Li, Wang, and Dai}]{zhu2020deformable}
Zhu, X.; Su, W.; Lu, L.; Li, B.; Wang, X.; and Dai, J. 2020.
\newblock Deformable DETR: Deformable Transformers for End-to-End Object Detection.
\newblock In \emph{International Conference on Learning Representations}.

\end{thebibliography}

\newpage

\setcounter{secnumdepth}{1}
\appendix

\section{More Implementation Details}
\label{sec:A}
We implement experiments of GeoBEV with 8 NVIDIA GeForce RTX 3090 GPUs. AdamW~\cite{loshchilov2017decoupled} with a decay weight of 0.01 is adopted as the optimizer and the learning rate is set to $2\times 10^{-4}$. We apply data augmentation in both image space and BEV space. The multi-view images are processed by random scaling with a range of $[0.86, 1.25]$
and horizontal flipping with a probability of 0.5. When training with CBGS~\cite{zhu2019class}, the range of random scaling is expanded to $[0.5, 1.25]$. Following BEVDet~\cite{huang2021bevdet}, the BEV features are processed by random scaling with a range of $[0.95, 1.05]$, random flipping of the X and Y axes with a probability of 0.5 and random rotating with a range of $[-22.5^{\circ}, 22.5^{\circ}]$. When processing images larger than or equal to $512\times 1408$, the size of BEV features is increased from $256\times 256$ to $512\times 512$.

\section{More Experiment Results}
\label{sec:B}

\subsection{More Results on RC-Sampling}

We provide more comparison results between RC-Sampling and the unoptimized Voxel-Sampling on time and memory costs in Tab.~\ref{tab:more_rcsampling}. Except for BEV size, the height of the voxel features also influences the efficiency of Voxel-Sampling. We partition the space by different numbers of heights and the experiment results show that higher height resolution leads to higher performance but also significantly increases the time and memory costs. On the contrary, RC-Sampling accumulates the features along the height before the sampling operations, which restores more complete height information than Voxel-Sampling by much less computation. From the experiment results, RC-Sampling achieves the same accuracy performance as Voxel-Sampling with a height resolution of 20 while spending less than 10\% of the time and memory costs. When producing the larger BEV representation, the time and memory costs of Voxel-Sampling become impracticable, while RC-Sampling remains deployment-friendly. Compared with BEVPoolv2~\cite{huang2022bevpoolv2}, RC-Sampling shows advantages in efficiency and accuracy performance while slightly increasing the memory cost.

\subsection{More Results on In-Box label}

Tab.~\ref{tab:inbox_optimize} shows more detailed results of the ablation study on the In-Box label. By default, the depth scores in the background region, where GT boxes are not available, are still supervised by the LiDAR label. 
If all background pseudo-points are regarded as negative classes, the requirement for LiDAR points during the training process will be eliminated, thereby significantly reducing data acquisition costs.
Compared to the model that does not use any depth labels, only using In-Box label increases 1\% mAP and 1.2\% NDS. The results in Tab.~\ref{tab:inbox_optimize} also show the effectiveness of several optimizations applied in In-Box label, including ignoring the pseudo-points in occlusion, the background pseudo-points within GT boxes and the pseudo-points behind the background surface. It illustrates that the optimized In-Box label which is more in line with the real world can lead to more precise detection.

\begin{table}[t]
    \centering
    \small
    \setlength{\tabcolsep}{0.5mm}{
    \begin{tabular}{l|c|c|cc|cc}
        \hline
        Method & BEV Size & H & mAP$\uparrow$ & NDS$\uparrow$ & Latency & Memory \\ 
        \hline
        \multirow{4}{*}{Voxel-Sampling} & $128\times 128$ & 5 & 0.337 & 0.458 & 1.77ms & 204.1MB\\
        & $128\times 128$ & 10 & 0.343 & 0.460 & 2.84ms & 252.1MB\\
        & $128\times 128$ & 20 & 0.342 & 0.464 & 3.18ms & 352.1MB\\
        & $256\times 256$ & 20 & 0.354 & 0.484 & 11.97ms & 952.1MB\\
        \hline
        \multirow{2}{*}{BEVPoolv2} & $128\times 128$ & - & 0.337 & 0.456 & 0.43ms & 19.5MB\\
        & $256\times 256$ & - & 0.344 & 0.474  & 0.62ms & 48.7MB \\
        \hline
        \multirow{2}{*}{RC-Sampling} & $128\times 128$ & - & 0.344 & 0.465 & 0.29ms & 20.0MB\\
        & $256\times 256$ & - & 0.358 & 0.482 & 0.53ms & 49.5MB \\
        \hline
        
    \end{tabular}
    }
    \caption{More comparison between different feature transformation modules. ``H'' denotes the height of the voxel features created by Voxel-Sampling. ``Latency'' and ``Memory'' are cost only by feature transformation approaches.}
    \label{tab:more_rcsampling}
\end{table}

\begin{table}
    \centering
    \small
    \setlength{\tabcolsep}{2mm}{
    \begin{tabular}{l|cccc|cc}
        \hline
        Label & w/o LiDAR & $O_A$ & $O_B$ & $O_C$ & mAP$\uparrow$ & NDS$\uparrow$\\ 
        \hline
        None & & & & & 0.330 & 0.454\\
        \hline
        \multirow{5}{*}{In-Box} & & & & & 0.351 & 0.470 \\
        & \checkmark & & & & 0.340 & 0.466 \\
        & & \checkmark & & & 0.353 & 0.472\\
        & & \checkmark & \checkmark & & 0.353 & 0.473 \\
        & & \checkmark & \checkmark & \checkmark & \textbf{0.356} & \textbf{0.474} \\
        \hline
    \end{tabular}}
    \caption{More ablation study on the In-Box label. ``w/o LiDAR'' denotes not using LiDAR points during training. ``$O_A$'' means ignoring the pseudo-points in occlusion. ``$O_B$'' means ignoring the background pseudo-points within GT boxes. ``$O_C$'' means ignoring the pseudo-points behind the background surface.}
    
    \label{tab:inbox_optimize}
\end{table}

\begin{figure*}[t]
    \centering
    \includegraphics[width=\linewidth]{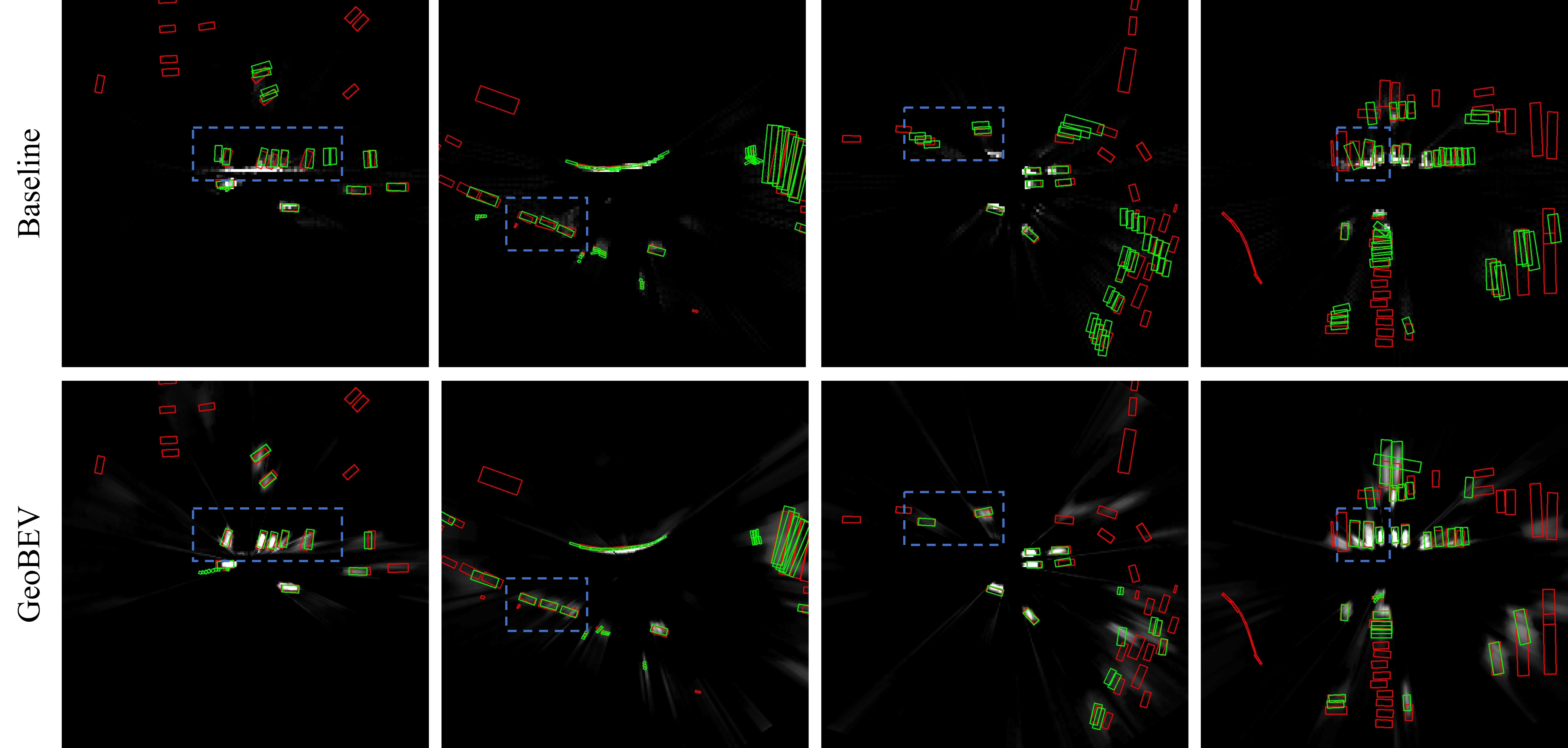}
    \caption{Visualization results on BEV representation of GeoBEV and its baseline. The background is filtered out to show the difference in the foreground. The red boxes and green boxes represent the ground truth and the predicted boxes, respectively. The dashed blue rectangles illustrate that the geometry guided BEV representations result in higher detection accuracy.}
    \label{fig:visualization}
\end{figure*}

\section{Visualization}
\label{sec:C}
We present the visualization results of GeoBEV and its baseline in Fig.~\ref{fig:visualization}. The L1 norm of each position in BEV features is first calculated. After that,   brightness positively correlated with the L1 norm is assigned in the figure. To show the difference in the foreground, we employ a pretrained instance segmentation model to provide foreground masks on the images. These masks are used to filter out the background regions before transforming image features into BEV features.

It can be found that the BEV representation of GeoBEV effectively restores the fine-grained geometric structure of the objects. This helps the network more clearly identify the size and orientation of the objects as illustrated by dashed blue rectangles, leading to improved detection performance.
\end{document}